\documentclass[runningheads]{llncs}

\usepackage{eccv}

\usepackage{eccvabbrv}

\usepackage{graphicx}
\usepackage{booktabs}

\usepackage[accsupp]{axessibility}  

\usepackage{hyperref}

\usepackage{orcidlink}
\usepackage[skip=2pt]{caption}
\usepackage{enumitem}
\usepackage{threeparttable}
\usepackage{makecell}
\usepackage{multirow}
\usepackage{float}
\usepackage{marvosym}

\newcommand{\jk}[1]{{\color{black}#1}}
\newcommand{\lrj}[1]{{\color{black}#1}}
\newcommand{\lrjfinal}[1]{{\color{black}#1}}

\title{PRET: Planning with Directed Fidelity Trajectory for Vision and Language Navigation}
\titlerunning{PRET}

\author{Renjie Lu\inst{1}\orcidlink{0009-0001-7808-3497}  \and
Jingke Meng\inst{1}\textsuperscript{(\Letter)}\orcidlink{0000-0001-5437-3070}  \and
Wei-Shi Zheng\inst{1,2,3}\orcidlink{0000-0001-8327-0003}
}
\authorrunning{Lu et al.}

\institute{School of Computer Science and Engineering, Sun Yat-sen University,\\
Guangzhou, China\\
\email{lurj3@mail2.sysu.edu.cn},\email{mengjke@gmail.com}, \email{wszheng@ieee.org}
\and
Peng Cheng Laboratory, Shenzhen, China
\and
Key Laboratory of Machine Intelligence and Advanced Computing,\\
Ministry of Education, Guangzhou, China
}

\begin{document}
\maketitle
\begin{abstract}
Vision and language navigation is a task that requires an agent to navigate according to a natural language instruction.
Recent methods predict sub-goals on constructed topology map at each step to enable long-term action planning. However, they suffer from high computational cost when attempting to support such high-level predictions with GCN-like models.
In this work, we propose an alternative method that facilitates navigation planning by considering the alignment between instructions and directed fidelity trajectories, which refers to a path from the initial node to the candidate locations on a directed graph without detours.
This planning strategy leads to an efficient model while achieving strong performance. Specifically, we introduce a directed graph to illustrate the explored area of the environment, emphasizing directionality. Then, we firstly define the trajectory representation as a sequence of directed edge features, which are extracted from the panorama based on the corresponding orientation. 
Ultimately, we assess and compare the alignment between instruction and different trajectories during navigation to determine the next navigation target. Our method outperforms previous SOTA method BEVBert on RxR dataset and is comparable on R2R dataset \lrjfinal{while largely reducing the computational cost.} Code is available: \href{https://github.com/iSEE-Laboratory/VLN-PRET}{\textcolor{blue}{https://github.com/iSEE-Laboratory/VLN-PRET}}.

\end{abstract}

\section{Introduction}
\label{sec:intro}

Enabling a robot to perform tasks on behalf of humans has been a longstanding objective in AI research. One such task is vision-and-language navigation (VLN)\cite{VLN,RxR,REVERIE,CVDN}, where an agent is required to navigate to a desired location by following natural language instructions provided by humans.
For example, given the instruction \textit{“Go up stairs and stop at the top in front of a mirror.”}, the agent needs to follow the instruction and stop at an appropriate location.
VLN has attracted numerous research interests\cite{speakerFollower,selfMonitoring,subInstruction,EnvDrop,PREVALENT,RecBERT,HAMT,SSM,DUET,AZHP,MetaExplore,BEVBert,GridMM}.

\begin{figure}[ht]
\centering
\includegraphics[width=0.8\linewidth]{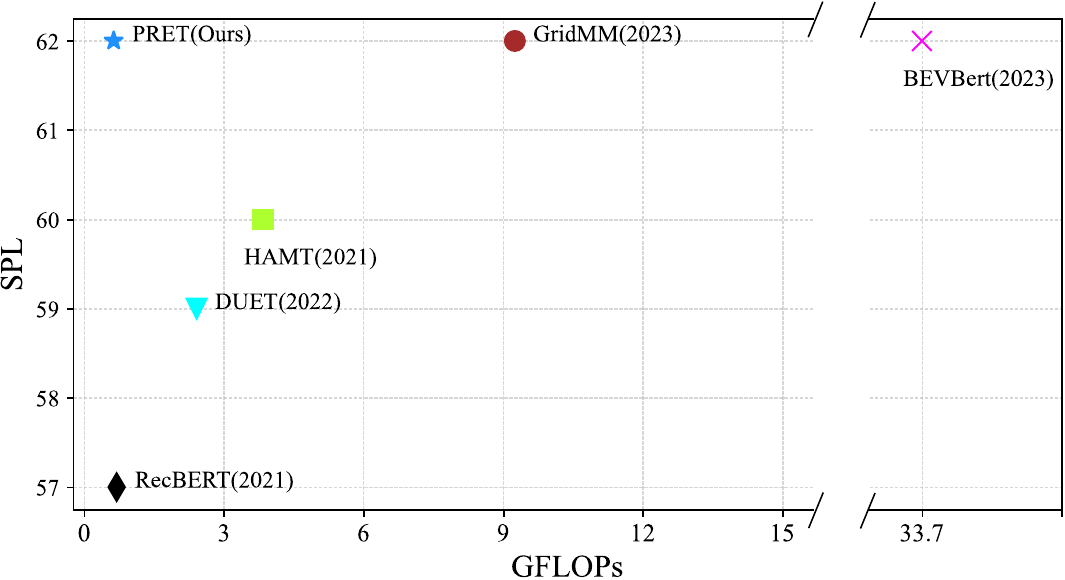}
\caption{Comparison of SPL\cite{SPL} and GFLOPs on R2R test unseen split dataset. Our method is comparable with previous SOTA methods while being more computational efficient. \lrjfinal{The computational cost of text encoder and visual encoder is omitted for fair comparison.}}
\label{fig:flops}
\vspace{-1em}
\end{figure}

Recent methods\cite{SSM,DUET,AZHP,MetaExplore,BEVBert} demonstrate the effectiveness of introducing maps to enlarge the decision space to improve planning strategies. Instead of predicting low-level actions limited to short movements, these methods construct a graph to keep track of all visited and navigable locations observed so far, which enables high-level planning by expanding the action space to encompass the entire explored area.
With the constructed topological maps, these methods predict actions in the global space via GCN-like models, where node features aggregate neighboring information.
However, repeatedly calculating the entire map to predict actions at every step, even if the topo-map has only minor changes, is inefficient. In addition, formulating vision information of an environment in graph structure is too coarse-grained for accurate decision-making. Previous methods\cite{DUET,BEVBert} address this problem by introducing an additional branch to incorporate fine-grained information, such as local image features and bird's-eye-view features. But this strategy increases the model complexity and further escalates computational cost.

\begin{figure}[t]
\centering
\includegraphics[width=0.7\linewidth]{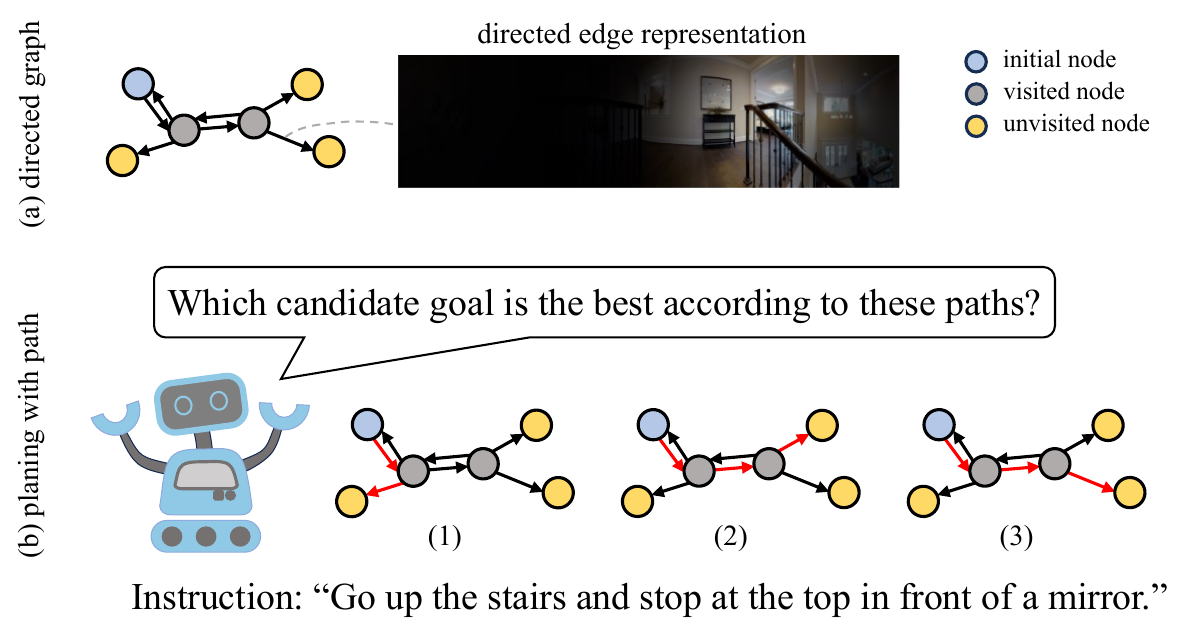}
\caption{Illustration of our approaches. (a) shows our directed graph representation. Each edge is assigned with an orientation-aware panorama feature. (b) depict our planning method. We select an unvisited node(colored yellow) to navigate towards next by choosing the fidelity path(colored red) that best aligned with instruction.
}
\label{fig:method}
\vspace{-1em}
\end{figure}

In this work, we present an alternative way that supports the global decision space while achieving comparable performance. Our method, named \textbf{P}lanning with Di\textbf{R}ected Fid\textbf{E}lity \textbf{T}rajectory(PRET), does not require calculating the entire graph at each step, nor does it rely on incorporating additional fine-grained information.
The main idea is to determine the next location to navigate by evaluating the alignment between the instruction and the visual observations along different trajectories between the start point to all the unvisited nodes, as shown in Figure \ref{fig:method}(b).
Specifically, we maintain a directed fidelity trajectory(colored in \textcolor{red}{red}) for each unvisited node. The fidelity trajectory refers to a path from initial node to an unvisited node without detours. We assess the alignment between each trajectory and the instruction, and select the unvisited node with the highest alignment score to navigate.
This planning strategy is efficient as we only need to compute for newly observed nodes. We estimate the instruction-trajectory alignment with transformer and further reduce the computational cost by utilizing the KV-cache\cite{kv-cache} technique. It also enables us to make decisions by taking into full consideration of the instruction-trajectory alignment.

In order to enhance the alignment between instructions and navigation trajectories, we propose to incorporate directionality in path representation.
Due to the inherent directional nature of the navigation process, it is crucial that the representation of trajectories in opposite directions is asymmetric, as visual observations are linked to observed orientation. In this work, we introduce the directed graph to depict the explored area of the environment and \textbf{firstly} define the trajectory representation as a sequence of directed edge features. These directed edge features are derived from the panorama based on the corresponding orientation (as depicted in \cref{fig:method}(a)).
By incorporating directionality in the graph edge features, we eliminate irrelevant panorama information and obtain a more accurate representation of the directed trajectory. 
Our method overcome existing methods\cite{DUET,AZHP,BEVBert} that commonly adopt the undirected graph to store panorama features on graph nodes, resulting in direction-invariant path representations, which fails to capture the important distinction between paths that traverse the same node from different directions. For instance, consider a scenario where agents pass through the same node from different directions. Despite the fundamentally different spatial contexts of these paths, the node representation remains unchanged.

We conduct experiments on R2R, RxR datasets to evaluate the efficacy of our proposed methods. The results demonstrate that PRET achieves strong performance while being more efficient than previous methods. Specifically, PRET achieves comparable performance compared to previous state-of-the-art method BEVBert\cite{BEVBert} on R2R dataset with only \lrjfinal{3\%} computational cost as shown in \cref{fig:flops}. On the RxR dataset, PRET outperforms previous methods and achieves the new state-of-the-art performance. Qualitative visualization also shows that our simplified model is able to learn complex backtracking strategies.

\section{Related Work}
\label{sec:related work}

\textbf{Vision-and-Language Navigation(VLN).}
Early VLN methods use sequence-to-sequence LSTMs with attention mechanism to encode language features and predict local actions. SpeakerFollower\cite{speakerFollower} first introduces back translation\cite{backTranslation} technique to ease the data scarcity problem in VLN. EnvDrop\cite{EnvDrop} proposes to augment environment by using dropout on feature space to avoid overfitting on visual input. Following this work, many environment augmentation approaches\cite{EnvMix,EnvEdit} is proposed. Some auxiliary tasks is proposed by \cite{selfMonitoring,AuxRN} for better guidance. More recently, transformer is explored how to be adopted in VLN. PRESS\cite{press} uses BERT\cite{bert} as the text encoder. PREVALENT\cite{PREVALENT} first propose to pretrain the transformer in VLN dataset, but still use it as encoder. More recent works\cite{RecBERT,HAMT,MTVM} explore how to memorize navigation history and use transformer to learn strong planning strategy.

\noindent\textbf{Navigation Strategy in VLN.}
Navigation strategy plays a key role in Vision-and-Language Navigation, as VLN requires agents to navigate in unseen environments and thus needs to explore and familiarize the environment. Works like \cite{regretful} investigate designing regretful agents to enable explicit exploration. Reinforcement learning methods \cite{RCM,EnvDrop} have also been explored to enhance navigation strategies. SSM\cite{SSM} constructs directed graphs to represent explored areas, with vision features on nodes and orientation on edges. DUET \cite{DUET} employs a dual-scale transformer to make local and global predictions. AZHP \cite{AZHP} constructs hierarchy graphs to facilitate exploration. MetaExplore \cite{MetaExplore} explicitly predicts whether to backtrack using a separate module.

\noindent\textbf{Maps For Navigation.}
Navigation research has a long history of using SLAM \cite{SLAM} to construct maps\cite{MetricMap,NeuralSLAM,SemanticMap} for planning.
In the VLN literature, several methods\cite{SSM,DUET} adopt topological graphs for global planning. However, the visual representation on topological maps is too coarse-grained for decision-making. To address this limitation, \cite{BEVBert,GridMM} introduce metric maps in VLN. While grid-based metric maps can precisely represent spatial layouts, they result in high computational costs. To balance representation ability and computational cost, BEVBert\cite{BEVBert} utilizes learnable hybrid topological and metric maps. However, the computational cost remains high.
\lrj{In our work, we propose a novel approach that involves constructing a directed topo-map and planning with trajectory. Our method is more efficient as we perform planning with trajectory instead of directly relying on GCN-like models to encode the entire map for planning. This also allows us to incrementally calculate embeddings for new trajectories and further reduces computational cost.}

\section{Method}
\label{sec:method}

\subsection{Problem Formulation}
In VLN with discrete environments\cite{VLN,RxR,CVDN,REVERIE}, an environment is an undirected graph $\mathcal G = \{\mathcal V ,\mathcal E\}$, where $\mathcal V=\{V_i\}_{i=1}^N$ represents $N$ navigable nodes and $\mathcal{E}$ denotes navigable edges.
At the beginning of navigation, an agent is initialized at a starting node and given a natural language instruction $W=\{w_i\}_{i=1}^L$ with $L$ words. The agent is required to interpret this instruction to navigate to the target location.

At step $t$ on node $V_t$, the agent observes 
(1) a panorama $\mathcal R_t=\{r_{t,i}\}_{i=1}^K$ represented by $K$ images from different views, $r_{t,i}$ is the extracted image feature of the $i$th view; 
(2) neighboring navigable nodes $\mathcal N(V_t)$, where $\mathcal N(V_t) \subset \mathcal V$; 
(3) orientations and coordinates of these neighboring nodes.
The agent should choose a neighboring node to step to or stop at current location. Navigation is considered successful if the agent stops within 3 meters of the target.

We treat navigation as a process of searching temporary target to navigate towards next among unvisited nodes on $\mathcal G_t$. A temporary target can be either a local neighboring node or remote unvisited node. By adding a virtual stop node connected to all nodes on the graph, we can also model the stop action.
We refer fidelity trajectory to a path from initial node to an unvisited node \textit{without detours}.

\subsection{Model Overview}

The overall framework is shown in \cref{fig:model}(a). Our planning method consists of three components:
(1) We construct a directed graph for explored area. Considering the directional nature of navigation, we propose an Orientation-aware Panorama Encoder(OPE) to extract orientation-aware vision features for edges. Edge sequence is used to represent directed trajectory and align with the instruction.
(2) We maintain a path \jk{(\ie directed fidelity trajectory)} to each unvisited node and assess the alignment between each path and the instruction at each step. Only few neighboring nodes are added at each step, we only compute for these relevant new paths. Since each node corresponds to a path, we calculate a path embedding with a matching assessment module(MAM) that encodes the instruction-path alignment and stores the embedding on the node.
(3) We compare the path embeddings of all unvisited nodes(candidates) with candidate comparison module(CCM)(shown in \cref{fig:model}(c)) to determine which aligns best with the instruction. We select the best aligned candidate node as the temporary target. Then the agent can navigate to the temporary target along the shortest path on constructed graph. Since the path embedding is an estimate of the alignment between instruction and a path to the node, CCM ensures that the agent navigates by fully considering the instruction-trajectory alignment.

\subsection{Orientation-Aware Directed Graph Construction}

We construct a directed graph to support path representation, incrementally adding nodes and edges as exploration proceeds.
Previous methods \cite{SSM,DUET,BEVBert} represent visited nodes with panorama features. However, panoramas of unvisited nodes are inaccessible as the agent has not visited them. So they represent these nodes with views towards them, leading to inconsistent representation between visited and unvisited nodes.
In contrast, we extract visual features for the directed edges on the graph. These edge features represent the visual observations when facing specific directions from each node. As edges rely only on observable views rather than node panoramas, they enable consistent representation for both visited and unvisited nodes.

\begin{figure*}[t]
\centering
\includegraphics[width=\linewidth]{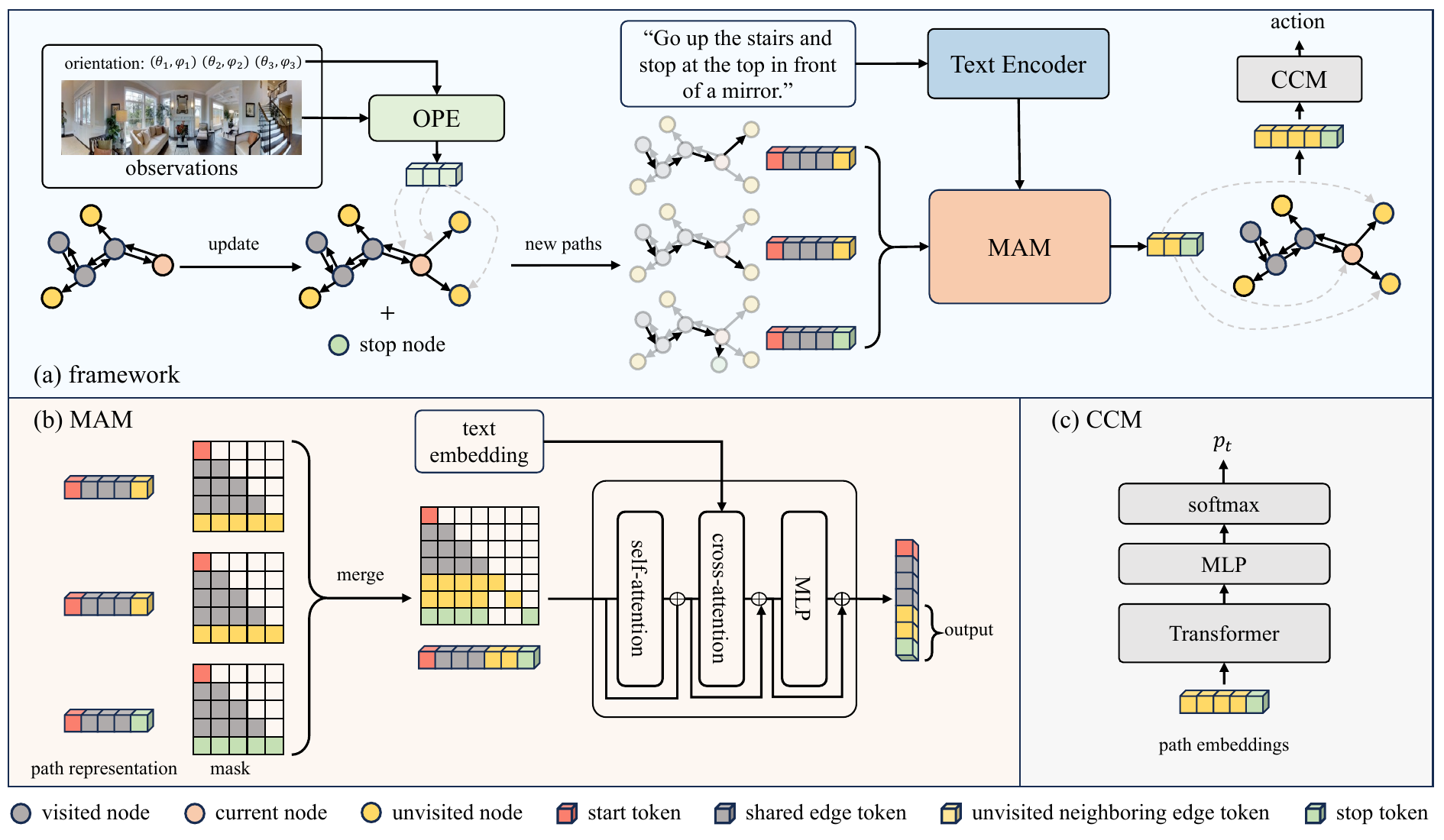}
\caption{\lrj{Illustration of our model. (a) is the overall framework of our method. At each step, we update the graph, extract path embeddings, and predict actions. (b) depicts the matching assessment module(MAM). Each token is an edge feature. We compute path embeddings for each newly observed nodes with cross-modal transformer and impose a causal mask to reduce computational cost. (c) shows the candidate comparison module(CCM). We gather path embeddings of unvisited nodes and forward them into a single layer transformer followed by a MLP to predict temporary target.
}}
\label{fig:model}
\end{figure*}

Let $\mathcal G_t = \{\mathcal V_t, \mathcal E_t\}$ be the directed graph at step $t$. Neighboring nodes $\mathcal N(V_t)$ are observed, and some nodes in $\mathcal N(V_t)$ have already existed in $\mathcal G_t$ while others are newly observed and added to $\mathcal G_t$. Besides, directed edges from $V_t$ to nodes in $\mathcal N(V_t)$ together with features extracted by orientation-aware panorama encoder(OPE) will also be added in $G_t$, as shown in \cref{fig:model}(a).

\subsubsection{Visual Representation For Edges.}
We extract orientation-aware panorama feature by using relative orientation as query to attend panorama features in cross-attention manner, \lrj{as shown in the OPE module in \cref{fig:model}(a)}. Assume that $(\phi, \theta)$ represents the relative heading and elevation of a specific edge when compared to the agent's current orientation. To extract feature for the edge, we first encode $(\phi, \theta)$ as follows:
\begin{equation}
x^a=
\begin{bmatrix}
\sin(\phi),\cos(\phi),\sin(\theta),\cos(\theta)
\end{bmatrix}W^a,
\end{equation}
where $W^a\in \mathbb R^{4\times d}$ are learnable weights and $d$ is the dimension of feature. Then we denote $X_t^a=\{x_{t,i}^a\}_{i=1}^{|\mathcal N(V_t)|}$ as orientation features of adjacent edges of $V_t$.

We then encode the panorama feature. As the panorama $\mathcal R_t=\{r_{t,i}\}_{i=1}^K$ is represented by $K$ images from different views and each view corresponds to a fixed orientation, we can also get the relative heading and elevation of each view. To encode the panorama, we first concatenate each view feature $r_{t,i}$ with corresponding relative orientation $(\phi_{t,i},\theta_{t,i})$ and then map it to dimension $d$ via a linear projection:
\begin{equation}
x_{t,i}^p=[r_{t,i};\sin(\phi_{t,i}),\cos(\phi_{t,i}),\sin(\theta_{t,i}),\cos(\theta_{t,i})]W^p.
\end{equation}
Then we denote $X_t^p=\{x_{t,i}^p\}_{i=1}^K$ as the panoramic view representation. By considering views as patches of the panorama and orientations as position embedding, $X_t^p$ is like the encoded ViT\cite{ViT} input.

We adopt transformer decoder\cite{transformer} to extract orientation-aware panorama feature:
\begin{equation}
\begin{aligned}
    E_t &= {\rm TransformerDecoder}(X_t^a, X_t^p),\\
\end{aligned}
\end{equation}
where $E_t=\{e_{t,i}\}_{i=1}^{|\mathcal N(V_t)|}$ are extracted edge features.
By taking $X_t^a$ as the query input and $X_t^p$ as the key-value input, panorama is attended by orientation in the first layer. In the subsequent layers, both orientation and vision information are used to query relevant views in the panorama. Therefore, each directed edge only focuses on a specific region of the panorama, thus orientation-aware panorama feature is extracted.

\subsection{Planning with Fidelity Trajectory}

We maintain a directed fidelity trajectory with a stack during navigation for each unvisited node on $\mathcal G_t$. When a new node is added, it is pushed into the stack. When the agent needs to backtrack, the top node is popped off the stack until a different node than the top is reached. In this way, detours are removed by popping nodes off the stack.
With the maintained trajectory of each node, we propose MAM to assess it's alignment with instruction. We then determine an unvisited node on $\mathcal G_t$ to navigate next by comparing this alignment with CCM.

\subsubsection{Matching Assessment Module(MAM).}
MAM is a multi-layer transformer decoder that extracts a path embedding for each path, which estimates how much the trajectory match with the instruction. These path embeddings are then stored on the corresponding nodes in graph $\mathcal{G}_t$.

For a single path with length $l$, we represent the path with the edge features along it, denoted as $X^h=[e_1, e_2, \cdots, e_l]$. $\{e_i\}_{i=1}^l$ are the edge features on the path. Then we add it with position embedding and forward it together with text embedding $X^w$ into MAM like follows:
\begin{equation}
\begin{aligned}
    X^{h\prime} &= X^h + P_l, \\
    X^o &= {\rm TransformerDecoder}(X^{h\prime}, X^w),
\end{aligned}
\end{equation}
where $P_l$ is a matrix including $l$ position embeddings\cite{transformer} and $X^w$ is the text embedding extracted by a text encoder, which is a multi-layer transformer\cite{transformer}. By utilizing two sequences from two modalities as inputs, the transformer is capable of effectively estimating their alignment, as extensively demonstrated in Vision and Language Pretraining studies\cite{UNITER,VILT,ALBEF}. The last token of the encoded sequence $X^o$ provides the path embedding, which is an estimation of the text-path alignment.

To support STOP action, we add a stop node that connected to all nodes in $\mathcal G_t$. Since we represent a path using edge features, we assign a learnable stop token, \lrj{initialized with a zero vector}, to all edges that point to the stop node. The path embedding corresponds to the stop node is stored on current node. So the agent decides to stop at current location when it find that the path up to here appending with stop token matches instruction the best. Here stop token serves as the End-of-Sentence token in text generation. For each path, we also add a learnable start token that serves as the Start-of-Sentence token.

At step $t$, considering the stop node, assume there are $N$ neighboring unvisited nodes to be updated. Computing path embeddings separately for these nodes requires multiple forward passes, which is inefficient. \textit{Actually, we can compute these embeddings in a single forward pass}, as shown in \cref{fig:model}(b). Note that these paths share the same prefix from the initial node to the current node. We can avoid duplicated computation by imposing a causal mask in the self-attention block. The causal mask is a lower triangular matrix that only allows a token to attend to its previous tokens. We merge these causal masks together and forward all relevant tokens into the transformer block. The merged mask still satisfies the constraint that each token can only attend to its previous tokens. We further reduce the computational cost by adopting the KV-cache technique\cite{kv-cache}, which stores the shared prefix tokens to avoid unnecessary re-calculation. In this way, we can equivalently compute the path embeddings for multiple newly observed nodes in a single forward pass. The last several output tokens(colored yellow and green) are path embeddings and used to update the graph, \ie, stored on corresponding graph nodes.

\subsubsection{Candidate Comparison Module(CCM).}
CCM compares the path embeddings from candidate nodes to predict a temporary target. As each path embedding encodes the alignment between a trajectory and the instruction, the CCM actually selects a path that is most aligned with the instruction. In this way, we make navigation planning fully according to the instruction-trajectory alignment.
As shown in \cref{fig:model}(c), we collect path embeddings $X_t^e$ and input them into CCM. CCM consists of a single-layer transformer encoder\cite{transformer} and a MLP. The transformer \jk{encoder} is responsible for \jk{the comparison of the candidates} and the MLP maps the encoded path embedding to a 1D score:
\begin{equation}
\begin{aligned}
    X_{t}^{e\prime} &= {\rm TransformerLayer}(X_t^e), \\
    s_t &= {\rm MLP}(X_t^{e\prime}), \\
    p_t &= {\rm softmax}(s_t).
\end{aligned}
\end{equation}
Score $s_t$ reflects the relative text-path alignment after comparison. An alternative way is removing the transformer and directly compute the alignment score $s_t$.  However, this method assesses each path independently without any comparison, which can make the decision-making process more challenging and potentially result in a decrease in performance(demonstrated in \cref{sec:experiments}).
It is then normalized with $\rm softmax$ and we select the highest score node as the temporary target. The agent will navigate to it along the shortest path on $\mathcal G_t$. If the stop node is selected, the agent believes the instruction is completed and stops the navigation.

\subsection{Pretraining and Fine-tuning}
\subsubsection{Pretraining.}
Pretraining is proved to be helpful in previous works\cite{PREVALENT,VLNBERT,HAMT,BEVBert}. PREVALENT\cite{PREVALENT} synthesized a substantial amount of data for pretraining, and we utilized the same data as their work.
We use Masked Language Modeling(MLM)\cite{bert} for pretraining. Concretely, we randomly replace 15\% of the input tokens with a special [MASK] token and forward these masked tokens into the text encoder. We also extract a sequence of edge features to represent the path. Then the encoded text tokens and path representation are input into an additional transformer decoder. The corresponding outputs are used to predict the masked tokens. MLM helps the model learn aligned text and path representations. In VLN, a key difference from other vision-and-language tasks is that we align a text sequence with a sequence of vision representations rather than a single image.

\subsubsection{Fine-tuning.}
We train the agent with a mixture of teacher-forcing and student-forcing as previous methods\cite{DUET,BEVBert}.
In the teacher-forcing stage, the agent navigate according the ground truth actions $a_t^*$ and the loss is calculated as follows:
\begin{equation}
    \mathcal L_{TF} = \frac{1}{T_{TF}}\sum\limits_{t=1}^{T_{TF}}{\rm CE}(p_t, a_t^*),
\end{equation}
$\rm CE$ is the cross entropy loss and $T_{TF}$ is the number of navigate steps.
In the student-forcing stage, we sample an action from the distribution predicted by the agent so that the agent can explore the environment and reduce the exposure bias. We train the agent with heuristic pseudo label $a_t^{pseudo}$:
\begin{equation}
    \mathcal L_{SF} = \frac{1}{T_{SF}}\sum\limits_{t=1}^{T_{SF}}{\rm CE}(p_t, a_t^{pseudo}).
\end{equation}
When the agent deviate from ground truth path, we take the nearest node on ground truth path as the pseudo label to encourage agent to learn backtracking strategy. If there is no unvisited node on ground truth path, we take the nearest node on the shortest path from current node to target node as the pseudo label.
The agent navigate two times to compute $\mathcal L_{TF}$ and $\mathcal L_{SF}$.
The total loss is the weighted sum of them:
\begin{equation}
    \mathcal L = \lambda \mathcal L_{TF} + (1-\lambda) \mathcal L_{SF},
\end{equation}
where $\lambda\in (0, 1)$ is the weight.

\section{Experiments}
\label{sec:experiments}
\subsection{Datasets and Evaluation Metrics}

\subsubsection{R2R.}
Room-to-Room (R2R) dataset\cite{VLN} contains 7,189 trajectories, with 3 instructions per trajectory, across 90 scenes. These scenes are divided into train, val unseen and test unseen splits with 61, 11 and 18 scenes respectively. A val seen split with the same scenes as the train split is also provided. All paths in R2R are the shortest paths between the start and target nodes.

\subsubsection{RxR.}
Room-across-Room(RxR)\cite{RxR} is a large multilingual dataset. The dataset has 126,000 instructions total, with 42,000 instructions in each of the three languages: English, Hindi, and Telugu. The paths in RxR are longer than those in R2R and are not the shortest possible routes. Additionally, the instructions in RxR are longer and contain more detailed descriptions compared to R2R.

\subsubsection{Evaluation Metrics.} 
We adopt the following evaluation metrics on R2R:
(1) Trajectory length (TL): the average length of the agent's path in meters. 
(2) Navigation error (NE): the average distance between the agent's final position and the target location.
(3) Success rate (SR): the ratio of successful navigations, where $NE < 3m$ is considered successful.
(4) SR penalized by path length (SPL)\cite{SPL}: TL longer than the shortest path is penalized.
As paths in RxR dataset are shortest paths, TL and SPL are unsuitable. Instead, on RxR we use nDTW and sDTW \cite{nDTW} to measure trajectory similarity between the agent and ground truth.
Please note that the fidelity trajectory that is removed detours is solely used for planning. It is not the agent's actual trajectory used for evaluation.

\subsection{Implementation Details}

\subsubsection{Module architecture.}
Text backbone is a 6 layer transformer encoder initialized with pretrained ALBEF\cite{ALBEF}. For the multilingual RxR dataset, we use 12 layer mRoberta\cite{mRoberta} instead. The layer numbers of the OPE, MAM, and CCM are 2, 4, and 1 respectively. All the hidden layer size is 768. Image features is extracted by DINOv2\cite{DINOv2}. We also report results with CLIP-ViT-base\cite{CLIP} feature for fair comparison.

\subsubsection{Training details.}
On R2R dataset, we first pretrain PRET with learning rate 2e-5 and batch size 16 on a single 24G 4090 GPU for 100,000 iterations($\sim5$ hours). Then we fine-tune the model with learning rate 1e-5 and batch size 8 on 4090 for 100,000 iterations($\sim25$ hours). Optimizer is AdamW\cite{AdamW}. $\lambda$ is set to $0.2$. Augmented dataset\cite{PREVALENT} is used in both pretraning and fine-tuning. The best result is selected by SPL on val unseen split.
On RxR dataset, we also follow the pretrain and fine-tune paradigm. Due to the longer instruction and trajectory, we use a smaller batch size 4. Marky\cite{marky} is used as augmented data. The best result is selected by sDTW on val unseen split.

\subsection{Results}

\begin{table}[t]
\centering
\caption{Comparison with other methods on R2R dataset. \lrjfinal{SPL is considered as the primary evaluation metric.}
}
\begin{threeparttable}
\resizebox{0.95\linewidth}{!}{
\begin{tabular}{l|*{4}{c}|*{4}{c}|*{4}{c}}
    \hline
    \makecell[c]{\multirow{2}{*}{Methods}}
                            & \multicolumn{4}{c|}{Val Seen}
                            & \multicolumn{4}{c|}{Val Unseen}
                            & \multicolumn{4}{c }{Test Unseen} \\
    \cline{2-13}
     & TL & NE$\downarrow$ & SR$\uparrow$ & SPL$\uparrow$
     & TL & NE$\downarrow$ & SR$\uparrow$ & SPL$\uparrow$
     & TL & NE$\downarrow$ & SR$\uparrow$ & SPL$\uparrow$ \\
    \hline
    Seq2Seq-SF\cite{VLN}        & 11.33 & 6.01  & 39 & -
                                & 8.39  & 7.81  & 22 & -
                                & 8.13  & 7.85  & 28 & 18 \\
    Speaker-Follower\cite{speakerFollower}
                                & -     & 3.36  & 66 & -
                                & -     & 6.62  & 35 & -
                                & 14.82 & 6.62  & 35 & 28 \\
    RCM\cite{RCM}               
                                & 10.65 & 3.53  & 67 & -
                                & 11.46 & 6.09  & 43 & -
                                & 11.97 & 6.12  & 43 & 38 \\
    Regretful\cite{regretful}   
                                & -     & 3.23  & 69 & 63
                                & -     & 5.32  & 50 & 41
                                & -     & 5.69  & 56 & 40 \\
    EnvDrop\cite{EnvDrop}       
                                & 11.00 & 3.99  & 62 & 59
                                & 10.70 & 5.22  & 52 & 48
                                & 11.66 & 5.23  & 51 & 47 \\
    PREVALENT\cite{PREVALENT}
                                & 10.32 & 3.67 & 69 & 65
                                & 10.19 & 4.71 & 58 & 53
                                & 10.51 & 5.30 & 54 & 51 \\
    NvEM\cite{NvEM}             & 11.09 & 3.44 & 69 & 65
                                & 11.83 & 4.27 & 60 & 55
                                & 12.98 & 4.37 & 58 & 54 \\
    SSM\cite{SSM}               
                                & 14.70 & 3.10 & 71 & 62
                                & 20.70 & 4.32 & 62 & 45
                                & 20.40 & 4.57 & 61 & 46 \\ 
    RecBert\cite{RecBERT}       
                                & 11.13 & 2.90 & 72 & 68
                                & 12.01 & 3.93 & 63 & 57
                                & 12.35 & 4.09 & 63 & 57 \\
    HAMT\cite{HAMT}             
                                & 11.15 & 2.51 & 76 & 72
                                & 11.46 & 2.29 & 66 & 61
                                & 12.27 & 3.93 & 65 & 60 \\
    MTVM\cite{MTVM}             
                                & -     & 2.67 & 74 & 69
                                & -     & 3.73 & 66 & 59
                                & -     & 3.85 & 65 & 59 \\
    DUET\cite{DUET}             
                                & 12.32 & 2.28 & 79 & 73
                                & 13.94 & 3.31 & 72 & 60
                                & 14.73 & 3.65 & 69 & 59 \\
    AZHP\cite{AZHP}             
                                & -     & -    & -  & - 
                                & 14.05 & 3.15 & 72 & 61
                                & 14.95 & 3.52 & 71 & 60 \\
    Meta-Explore\cite{MetaExplore}  
                                & 11.95 & \textbf{2.11} & \textbf{81} & \textbf{75}
                                & 13.09 & 3.22 & 72 & 62
                                & 14.25 & 3.57 & 71 & 61 \\
    GridMM\cite{GridMM}         & -     & -    & -  & -
                                & 13.27 & 2.83 & 75 & 64
                                & 14.43 & 3.35 & 73 & 62 \\
    BEVBert\cite{BEVBert}       
                                & 13.56 & 2.17 & \textbf{81} & 74
                                & 14.55 & \textbf{2.81} & \textbf{75} & 64
                                & 15.87 & 3.13 & \textbf{73} & 62 \\
    \hline
    Ours(CLIP)                  & 11.48 & 2.60 & 74 & 69
                                & 12.21 & 3.12 & 71 & 63
                                & 13.87 & 3.12 & 72 & 62 \\
    Ours(DINOv2)                & 11.25 & 2.41 & 78 & 72
                                & 11.87 & 2.90 & 74 & \textbf{65}
                                & 12.21 & \textbf{3.09} & 72 & \textbf{64}   \\
    \hline
\end{tabular}
}
\end{threeparttable}
\label{tab:results_R2R}
\end{table}

\subsubsection{Comparison on R2R.}
\cref{tab:results_R2R} compares our approach with previous methods. 
On the val unseen and test unseen split, our method achieves comparable performance with previous SOTA method BEVBert on the primary metric SPL, while our method significantly reduces the computational cost. Besides, BEVBert\cite{BEVBert} adopts additional depth information to construct bird's-eye view input. On the val seen split, our method performs sub-optimal as PRET does not take the environment layout as input, which means it less tends to overfit on the seen environment. Also, in the VLN literature, all visual features are pre-extracted and remains unchanged, the fixed feature may not contains appropriate feature for navigation. Therefore, we also investigate different vision features for VLN, as shown in \cref{tab:results_R2R}. We find that DINOv2\cite{DINOv2} feature is better than CLIP feature. DINOv2 is trained on large-scale curated data in self-supervised manner, which leads to more general purpose feature. Moreover, DINOv2 feature contains geometric information as it can be used to predict depth map according to \cite{DINOv2}. Geometric information is shown helpful in \cite{GeoVLN}.

\begin{table}[ht]
\centering
\caption{Comparison of \lrjfinal{latency and computational cost} on R2R val unseen. * indicates we reproduce DUET with CLIP feature for fair comparison. When computing GFLOPs,
\lrjfinal{we simplify the calculation by assuming that each vertex of the graph is connected to 4 adjacent vertices (average degree in R2R), and at each step, 3 new vertices are encountered. The computational cost of text encoder and visual encoder is omitted for fair comparison.}}
\resizebox{0.85\linewidth}{!}{
\begin{tabular}{l|cccc|cc|ccc|c}
    \hline
    \multirow{2}{*}{Methods}
    & \multicolumn{4}{c|}{R2R val unseen}
    & \multicolumn{2}{c|}{Latency(ms)}
    & \multicolumn{3}{c|}{GFLOPs} & \multirow{2}{*}{Params} \\
    \cline{2-10}
    & TL & NE$\downarrow$ & SR$\uparrow$ & SPL$\uparrow$
    & train & inference
    & 1 step & 10 steps & 20 steps & \\
    \hline
    DUET*\cite{DUET}      & 14.0 & 3.2   & 71.6 & 61.1
                          & 11.8 & 8.6
                          & 2.4  & 4.8   & 7.5 & 90M \\
    BEVBert\cite{BEVBert} & 14.6 & \textbf{2.8}   & \textbf{74.9} & \textbf{63.6}
                          & 16.2 & 10.6
                          & 33.7 & 35.5  & 37.5 & 90M \\
    \hline
    Ours 
          & 12.2 & 3.1 & 71.0 & 62.7
          & \textbf{7.7} & \textbf{4.1}
          & \textbf{0.6}  & \textbf{0.9}  & \textbf{1.2}  & \textbf{64}M \\
    \hline
\end{tabular}
}
\label{tab:flops}
\end{table}

\cref{tab:flops} shows the \lrjfinal{parameter count, computational cost, and latency of different graph-based methods.} We present GFLOPs of different navigation steps.
Our method contains fewer parameters as we use a single stream model unlike \cite{DUET,BEVBert} use dual stream transformer. Besides, our \lrjfinal{decoder} computational cost is only 19\% of DUET and 3\% of BEVBert at step 10. These suggests that our method is significantly more efficient than previous SOTA graph-based methods. \lrjfinal{As shown in \cref{tab:flops}, this results in lower training and inference latency}.

\begin{table}[ht]
\centering
\caption{ Comparison with previous methods on RxR dataset. \dag indicates approaches that utilize additional augmented data. nDTW represents the normalized DTW distance between the agent's trajectory and ground truth path. sDTW is the nDTW weighted by success rate.
}
\setlength\tabcolsep{2pt}  
\resizebox{0.8\linewidth}{!}{
\begin{tabular}{l|*{4}{c}|*{4}{c}}
    \hline
    \makecell[c]{\multirow{2}{*}{Methods}} 
                            & \multicolumn{4}{c|}{Val Seen}
                            & \multicolumn{4}{c }{Val Unseen} \\
    \cline{2-9}
     & NE$\downarrow$ & SR$\uparrow$ & nDTW$\uparrow$ & sDTW$\uparrow$
     & NE$\downarrow$ & SR$\uparrow$ & nDTW$\uparrow$ & sDTW$\uparrow$ \\
    \hline
    LSTM\cite{RxR}              & 10.7 & 25.2 & 42.2 & 20.7
                                & 10.9 & 22.8 & 38.9 & 18.2
                                \\
    EnvDrop+\cite{EnvDrop+}     & -    & -    & -    & -
                                & -    & 43.6 & 55.7 & -    
                                \\
    HAMT\cite{HAMT}             & -    & 59.4 & 65.3 & 50.9
                                & -    & 56.5 & 63.1 & 48.3
                                \\
    EnvEdit\cite{EnvEdit}       & -    & 67.2 & 71.1 & 58.5
                                & -    & 62.8 & 68.5 & 54.6
                                \\
    MPM\cite{MPM}               & -   & 67.7 & 71.0 & 58.9
                                & -   & 63.5 & 67.7 & 54.5
                                \\
    MARVEL\cite{marky}\dag      & 3.0 & 75.9 & 79.1 & 68.8
                                & 4.5 & 64.8 & 70.8 & 57.5
                                \\
    BEVBert\cite{BEVBert}\dag   & 3.2 & 75.0 & 76.3 & 66.7
                                & 4.0 & 68.5 & 69.6 & 58.6
                                \\
    \hline
    
    Ours(CLIP)\dag & 2.6 & 77.0 & 78.0 & 67.5
                   & 3.3 & 71.2 & 71.8 & 60.4 \\
    Ours(DINOv2)\dag & \textbf{2.4} & \textbf{79.3} & \textbf{80.4} & \textbf{70.7}
             & \textbf{3.2} & \textbf{72.8} & \textbf{73.4} &    \textbf{62.4}
            \\
    \hline
\end{tabular}
}
\label{tab:results_RxR}
\end{table}

\subsubsection{Comparision on RxR.}
\cref{tab:results_RxR} reports the results on the RxR dataset. RxR is challenging as instructions and paths in RxR are longer than R2R. While the alignment between path and detailed path description is what our method skills at, our method outperforms previous methods among all metrics at a lower computational cost. PRET achieves a 1.8\% improvement on the main metric sDTW on the val unseen split, demonstrating that our model better understands and follows instructions. This result highlights the advantage of our planning with trajectory strategy, which considers the alignment between instructions and paths and enables efficient decision-making in the global space. Additionally, with DINOv2 features, the result is further improved.

\subsection{Ablation Study}

\begin{table}
\begin{minipage}{0.42\textwidth}
\centering
\caption{Comparison of undirected and directed path representation.}

\resizebox{0.9\linewidth}{!}{
\begin{tabular}{l|*{4}{c}}
    \hline
    Methods & TL & NE$\downarrow$ & SR$\uparrow$ & SPL$\uparrow$  \\
    \hline
    undirected & 15.62 & 3.59 & 68.28 & 56.77 \\
    directed   & 11.87 & 2.90 & 73.78 & 65.16 \\
    \hline
\end{tabular}
}
\label{tab:ablation_graph}
\end{minipage}
\begin{minipage}{0.55\textwidth}
\centering
\caption{ Ablation study on modules. }
\resizebox{0.9\linewidth}{!}{
\begin{tabular}{ccl|*{4}{c}}
    \hline
    && Methods & TL & NE$\downarrow$ & SR$\uparrow$ & SPL$\uparrow$  \\
    \hline
    1 && MAM          & 12.04 & 3.99 & 62.32 & 54.48 \\
    2 && MAM+CCM      & 12.15 & 3.54 & 65.94 & 57.32 \\
    3 && MAM+OPE      & 12.18 & 3.15 & 71.60 & 63.07 \\
    4 && MAM+OPE+CCM  & 11.87 & 2.90 & 73.78 & 65.16 \\
    \hline
\end{tabular}
}
\label{tab:ablation_method}
\end{minipage}
\end{table}

We conducted ablation experiments on various components of PRET, including the directed path representation and module ablation. Results are reported on the R2R val unseen split.

\newpage
\subsubsection{Directionality in Path Representation.}
We compare whether to incorporate directionality for path representations in \cref{tab:ablation_graph}. Undirected path is represented by a sequence of node features. The node features is extracted by forwarding panoramic views into a 2-layer transformer encoder and averages the output tokens. Directed path is represented by a sequence of orientation-aware panorama features. According to the results, directed path representation outperforms undirected node features among all metrics. Specifically, it gains 8.39\% improvement on SPL and 5.5\% on SR. This indicates that edge feature is more suitable to represent the directional navigation process and align with the instruction. Node features does not distinguish different path directions and provide redundant information in a path, which hinders the alignment.

\begin{figure}[t]
\centering
\includegraphics[width=0.5\linewidth]{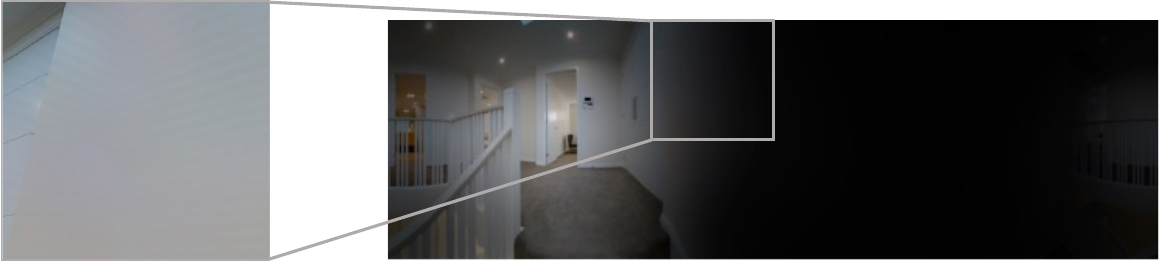}
\captionof{figure}{Comparison of orientation panoramic view and single candidate view.}
\label{fig:pano_view}
\end{figure}

\subsubsection{Module ablation.}

\cref{tab:ablation_method} shows ablation results on modules. Row 4 is the full model introduced in \cref{sec:method}. OPE extracts orientation-aware features for edges, MAM extracts path embeddings while CCM compares them to select the temporary target. When removing CCM (Row 3), we only predict a single alignment score rather than a path embedding for each path without comparison. The score is then normalized and used to predict the temporary goal. Performance drops 2.09\% in this case, as selecting a path based solely on a single score, without comparing different paths, is more difficult.

In row 2, we remove OPE, panorama features are not used and only a single 60-degree view towards each neighboring node is adopted as edge features. Performance drops 7.84\% as the limited field-of-view cannot fully represent the path. The trajectory represented by these views is discrete and does not form a continuous change of views. In some cases, a single view towards candidates provides no information for navigation, as shown in \cref{fig:pano_view}. When go up the stairs, the upper view only sees the wall and does not help the navigation at all. While using the panorama, the agent can dynamically see a larger region. These results shows our graph representation with panorama encoder provides rich representation for trajectories. Row 1 demonstrates the performance when using only MAM. The performance is poor without other model components, demonstrating the effectiveness of these modules.

\subsection{Visualization}
\vspace{-2em}
\begin{figure}[ht]
\centering
\includegraphics[width=0.95\linewidth]{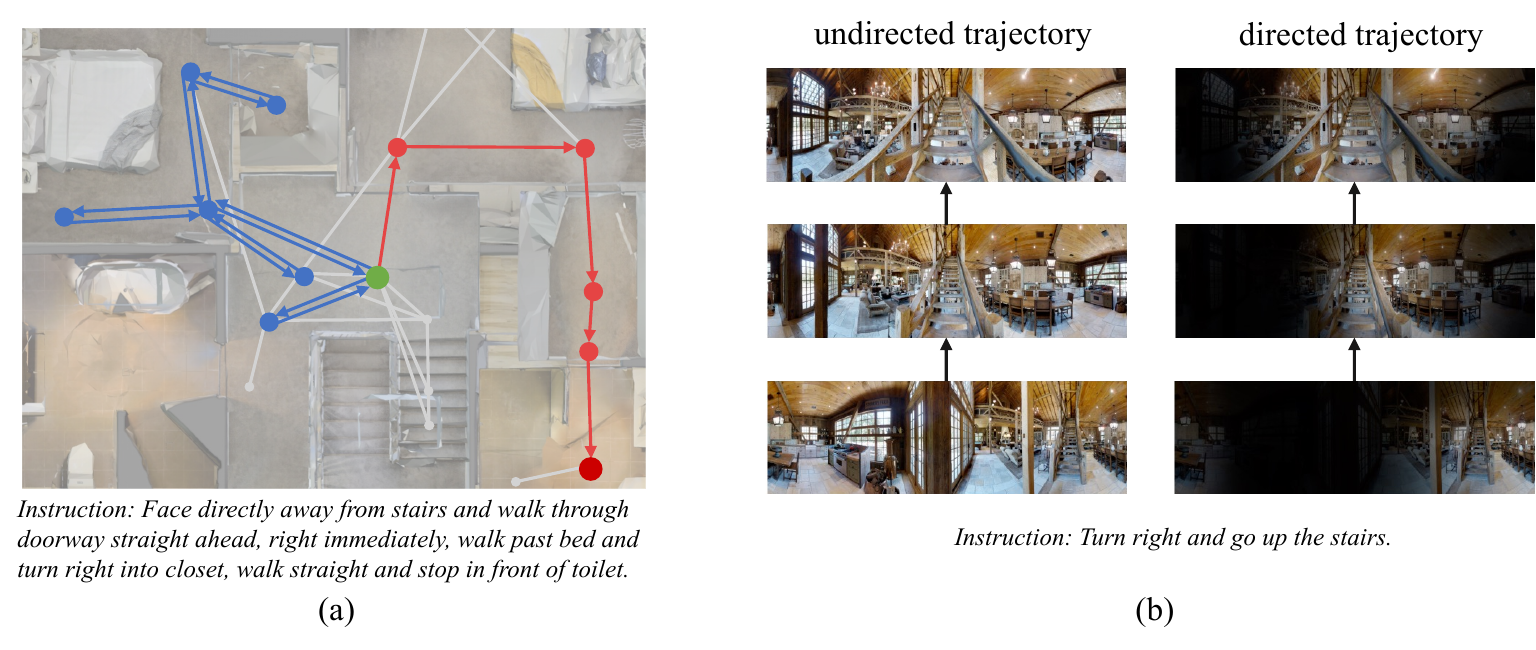}
\caption{
(a) Visualization of the agent's navigation process, showcasing its ability to learn a backtracking strategy.
(b) Visualizing attention weights in OPE to illustrate the distinction between undirected and directed trajectory representations.
}
\label{fig:visualization}
\end{figure}
\vspace{-1em}

\cref{fig:visualization}(a) shows a qualitative example of our model's behavior. Initially placed at the green starting node, the agent wrongly decides to explore the upper left area(blue path). It takes many unnecessary detours there, but through comparing the trajectory alignment with the instructions, realizes these paths do not match the text well. Therefore, it backtracks to the start and correctly navigates to the destination(red path).
This demonstrates the capability of PRET to learn complex backtracking behaviors.

\cref{fig:visualization}(b) shows the difference between undirected and directed trajectory representations. The former represents a path using panorama features stored on nodes, which eliminates directional discrepancies since the same node provides the same feature regardless of orientation, and contains redundant information compared to \textit{“stairs”} in the instruction. In contrast, our proposed approach represents a path using orientation-aware panorama features on edges. This eliminates redundant information and focuses on the \textit{“stairs”} better captures the directional navigation process.

\section{Conclusion}

In this work, we present an efficient method that planning with directed fidelity trajectory, explicitly leveraging the alignment between instructions and trajectories for navigation. Additionally, we consider the directional nature of navigation and introduce a directed graph construction during navigation, storing vision information on directed edges. This approach provides rich directed path representation and enhances instruction-trajectory alignment. Experiments demonstrate that our method is achieves strong performance while being significantly more efficient than previous methods. Our method has certain limitations that we need to address. For example, navigation requires environment layout information in some cases. We recognize the need to investigate potential solutions to overcome these challenges.

\section*{Acknowledgments}

This work was supported partially by the National Key Research and Development Program of China (2023YFA1008503), NSFC(U21A20471, 62206315), Guangdong NSF Project (No. 2023B1515040025, 2020B1515120085, 2024A1515-010101), Guangzhou Basic and Applied Basic Research Scheme(2024A04J4067).

%
%
\bibliographystyle{splncs04}
\bibliography{main}

\end{document}